\pdfoutput=1

\documentclass[11pt]{article}

\usepackage[final]{ACL2023}
\usepackage{times}
\usepackage{latexsym}

\usepackage[T1]{fontenc}

\usepackage[utf8]{inputenc}

\usepackage{microtype}

\usepackage{inconsolata}

\usepackage{graphicx}

\usepackage{algorithm}
\usepackage{algorithmic}
\usepackage{amsfonts}
\usepackage{amsmath}
\usepackage{booktabs}
\usepackage{multirow}
\usepackage{graphicx}
\usepackage{enumitem}
\usepackage{textcomp}
\usepackage{booktabs}
\usepackage{subfigure}
\usepackage{array}
\usepackage{bm}
\usepackage{tabularx}
\usepackage{tabulary}
\usepackage{array}
\usepackage{color,ulem}
\usepackage{float}  
\usepackage{placeins}

\title{CartesianMoE:  Boosting Knowledge Sharing among Experts\\ via Cartesian Product Routing in Mixture-of-Experts}

\author{
  Zhenpeng Su\textsuperscript{\rm 1,2}\quad Xing Wu\textsuperscript{\rm 1,2} \quad Zijia Lin\textsuperscript{\rm 3}\footnotemark[1]\quad Yizhe Xiong\textsuperscript{\rm 3}\quad \textbf{Minxuan Lv} \textsuperscript{\rm 1,2} \\ \quad \textbf{Guangyuan Ma}\textsuperscript{\rm 1,2} \quad \textbf{Hui Chen} \textsuperscript{\rm 3}\footnotemark[1] \quad \textbf{Songlin Hu}\textsuperscript{\rm 1,2}\footnotemark[1] \quad \textbf{Guiguang Ding}\textsuperscript{\rm 3}\\
  \textsuperscript{\rm 1}Institute of Information Engineering, Chinese Academy of Sciences\\
  \textsuperscript{\rm 2}University of Chinese Academy of Sciences \quad
  \textsuperscript{\rm 3}Tsinghua University \\
  \tt\small \texttt{\{suzhenpeng,wuxing,maguangyuan,lvminxuan,husonglin\}@iie.ac.cn}\\
  \tt\small \texttt{linzijia07@tsinghua.org.cn}
  \tt\small \texttt{\{huichen,dinggg\}@tsinghua.edu.cn, xiongyizhe2001@gmail.com}\\
}

\begin{document}
\maketitle
\renewcommand{\thefootnote}{\fnsymbol{footnote}} 
\footnotetext[1]{Corresponding authors.} 
\footnotetext{This work was supported by Beijing Natural Science Foundation (L247026) and National Natural Science Foundation of China (No 62441235).}
\renewcommand{\thefootnote}{\arabic{footnote}}
\begin{abstract}
Large language models (LLM) have been attracting much attention from the community recently, due to their remarkable performance in all kinds of downstream tasks. According to the well-known scaling law, scaling up a dense LLM enhances its capabilities, but also significantly increases the computational complexity. Mixture-of-Experts (MoE) models address that by allowing the model size to grow without substantially raising training or inference costs. Yet MoE models face challenges regarding knowledge sharing among experts, making their performance somehow sensitive to routing accuracy. 
To tackle that, previous works introduced shared experts and combined their outputs with those of the top $K$ routed experts in an ``addition'' manner. In this paper, inspired by collective matrix factorization to learn shared knowledge among data, we propose CartesianMoE, which implements more effective knowledge sharing among experts in more like a ``multiplication'' manner. 
Extensive experimental results indicate that CartesianMoE outperforms previous MoE models for building LLMs, in terms of both perplexity and downstream task performance. And we also find that CartesianMoE achieves better expert routing robustness.

\end{abstract}

\section{Introduction}

Large language models (LLM) have demonstrated impressive performance across various downstream natural language tasks~\cite{DBLP:journals/corr/abs-2302-13971,DBLP:conf/acl/Dai0MZSCW22,DBLP:conf/nips/BrownMRSKDNSSAA20,DBLP:journals/corr/abs-2312-11805,DBLP:journals/corr/abs-2204-02311,radford2019language,DBLP:journals/corr/abs-2112-11446,DBLP:conf/icml/BidermanSABOHKP23}. Moreover, the well-known scaling law suggests that, as the model size increases, the model capabilities will continue to improve~\cite{DBLP:journals/corr/abs-2001-08361,DBLP:journals/corr/abs-2203-15556}. However, for dense LLMs, the computational costs of scaling up their model sizes can become prohibitively high.
To tackle that, sparse activation networks are proposed~\cite{DBLP:journals/corr/abs-1904-10509,DBLP:conf/icml/DuHDTLXKZYFZFBZ22}. They reduce computational costs by activating only a subset of parameters for each input. A prominent approach among them is the mixture-of-experts (MoE)~\cite{DBLP:conf/iclr/LepikhinLXCFHKS21,DBLP:conf/icml/DuHDTLXKZYFZFBZ22,DBLP:journals/corr/abs-2401-06066,DBLP:journals/jmlr/FedusZS22,DBLP:conf/nips/RollerSSW21}, which involves training multiple experts but using only a subset to process each input, with each expert generally being a feed-forward network (FFN). Compared to dense LLMs of equivalent sizes, MoE LLMs effectively reduces computational costs while delivering comparable results, in terms of both perplexity (PPL) and downstream task performance~\cite{DBLP:conf/iclr/LepikhinLXCFHKS21,DBLP:conf/icml/DuHDTLXKZYFZFBZ22,DBLP:journals/corr/abs-2401-06066,su2024maskmoe,huang-etal-2024-harder,yang-etal-2024-xmoe,zhao-etal-2024-hypermoe}.

Conventional MoE models, like ~\cite{DBLP:conf/iclr/LepikhinLXCFHKS21,DBLP:journals/jmlr/FedusZS22,DBLP:conf/icml/DuHDTLXKZYFZFBZ22}, activate the top \( K \) routed experts among the total \( N \) experts. 
Due to the independent training of all experts, they rarely share learned knowledge, and thus routing fluctuations can affect the output substantially, making the performance of such MoE models somehow sensitive to the routing accuracy.   
To tackle that, ~\cite{DBLP:journals/corr/abs-2401-06066,DBLP:conf/icml/RajbhandariLYZA22} suggests using several fixed-activated shared experts to store shared knowledge, in addition to the top$K$ routed experts. And it has been well-validated to improve MoE model performance. 
With shared experts, ~\cite{DBLP:journals/corr/abs-2401-06066,yang-etal-2024-xmoe} further split full-sized experts into more fine-grained experts to enhance representation specialization and gain additional performance improvement, where the \( N \) experts are split into \( mN \) smaller ones, and the top \( mK \) routed ones of them are activated. 

\begin{figure*}
    \centering
    \scalebox{0.62}{
    \subfigure[Conventional Top-2 routing for full-sized experts]{
        \includegraphics[width=0.5\textwidth]{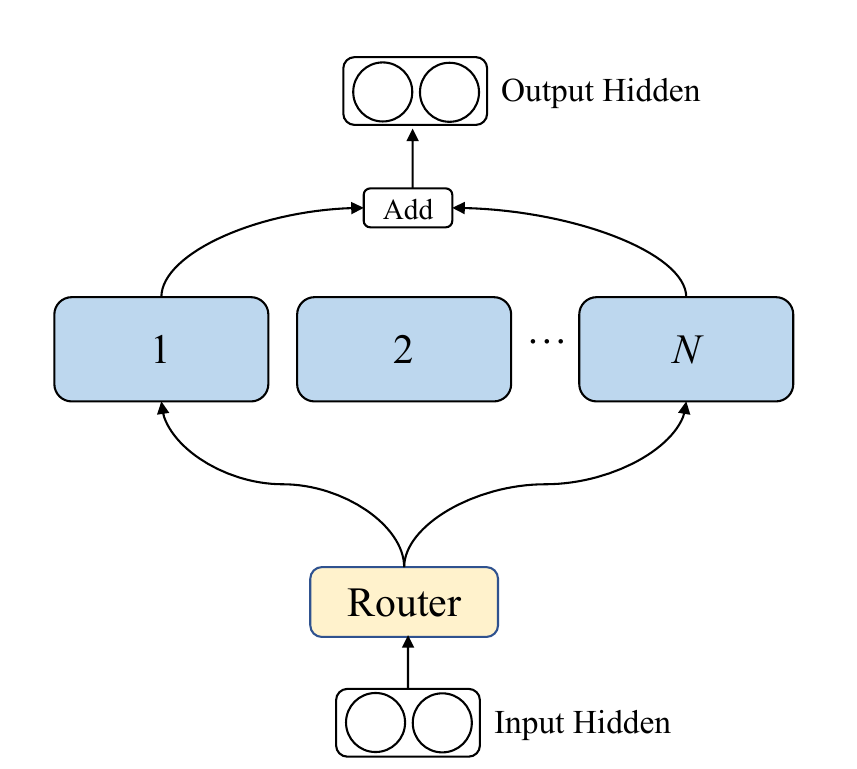}
        \label{fig:subfig1}
    } \hfill 
    \subfigure[Top-4 routing for half-sized fine-grained experts]{
        \includegraphics[width=0.5\textwidth]{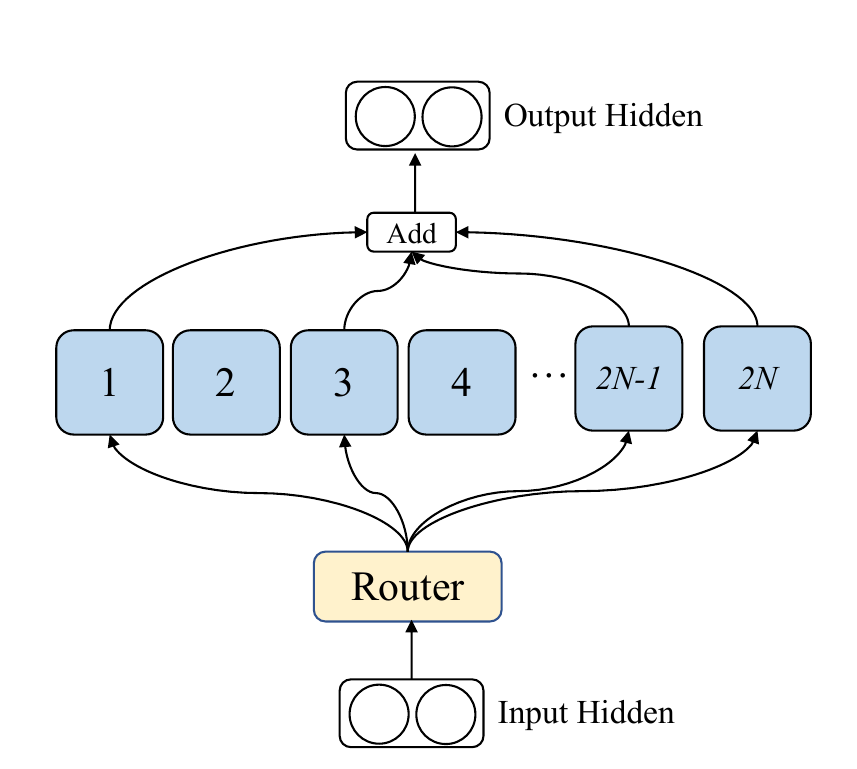}
        \label{fig:subfig2}
    } \hfill 
    \subfigure[Top-4 Cartesian Product routing in CartesianMoE]{
        \includegraphics[width=0.5\textwidth]{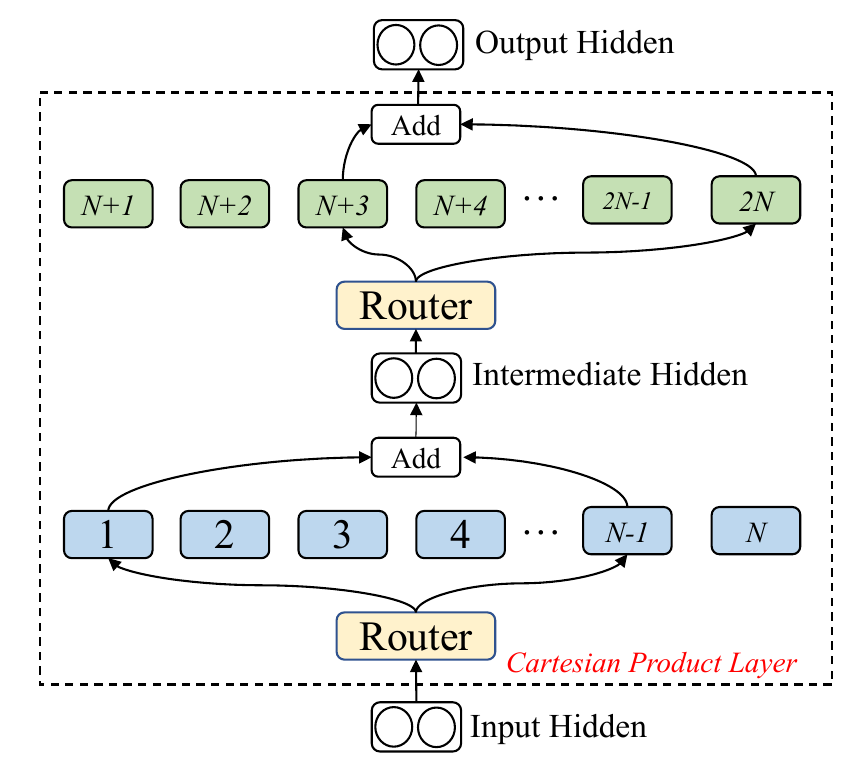}
        \label{fig:subfig3}
    }
    }
    \caption{Illustration of CartesianMoE. Subgraph a) represents the conventional top-2 routing for full-sized experts, subgraph b) illustrates the top-4 routing for half-sized fine-grained experts, and subgraph c) shows the top-4 Cartesian Product routing (i.e., top-2 routing for each sub-layer) in the proposed CartesianMoE. All subgraphs share the same numbers of model parameters and activated parameters.}
    \label{fig:overall_fig}
\end{figure*}

The remarkable shared-expert method essentially merges the shared knowledge (i.e., outputs of shared experts) with the specific knowledge (i.e., outputs of the routed experts) in an ``addition'' manner. For instance, with a shared expert FFN$_{a}$ and several routed experts FFN$_{b}$, FFN$_{c}$, and FFN$_{d}$, the knowledge sharing among experts can be represented as: FFN$_{a}$ + FFN$_{b}$, FFN$_{a}$ + FFN$_{c}$, and FFN$_{a}$ + FFN$_{d}$. Inspired by collective matrix factorization to learn shared knowledge among data~\cite{relation:CMF}, in this paper we propose to represent knowledge sharing among experts in an alternative ``multiplication'' manner, i.e., FFN$_{a}$ $\cdot$ FFN$_{b}$, FFN$_{a}$ $\cdot$ FFN$_{c}$, and FFN$_{a}$ $\cdot$ FFN$_{d}$. Specifically, by defining two sets of sub-experts $\{$FFN$_{a}^1$, FFN$_{b}^1$, $\ldots$$\}$ and $\{$FFN$_{a}^2$, FFN$_{b}^2$, $\ldots$$\}$, we derive each expert to be the combination of any two sub-experts from both sets respectively, like FFN$_{aa}$ = FFN$_{a}^1$ $\cdot$ FFN$_{a}^2$ or FFN$_{ab}$ = FFN$_{a}^1$ $\cdot$ FFN$_{b}^2$. In that sense, each expert share an identical sub-expert with many others. It can also be seen that, all the experts can be derived by the Cartesian product between both sub-expert sets, and thus we term our proposed method as CartesianMoE. Specifically, in our proposed CartesianMoE, we replace the conventional MoE layer as a \textit{Cartesian Product Layer}, which consists of two sequential MoE sub-layers, each denoting a set of sub-experts, as illustrated in Fig.~\ref{fig:overall_fig}. Then the routing process to select routed experts is also divided into the two MoE sub-layers, termed as Cartesian Product routing.

Extensive experiments on building MoE LLMs show that CartesianMoE yields superior performance than previous counterparts, using the same number of model parameters and activated parameters. 
CartesianMoE also shows better routing robustness. We argue that the superiority of CartesianMoE comes from its more fine-grained knowledge sharing among experts. Specifically, compared to the shared-expert method that requests all routed experts to always share the same global knowledge held by the fixed shared experts, CartesianMoE allows to divide experts into groups with each sharing some group-wise knowledge. In that sense, CartesianMoE is supposed to be also equipped with shared experts, so as to form a ``global shared knowledge + group-wise shared knowledge + expert-specific knowledge'' system.

Our contributions are summarized as follows:
\begin{itemize}[leftmargin=*]
    \item Inspired by collective matrix factorization to learn shared knowledge among data, we analyze the feasibility of enabling knowledge sharing among experts in a ``multiplication'' manner, an alternative to the ``addition'' manner proposed by the shared-expert method.
    \item We propose CartesianMoE, which derives experts via the Cartesian Product of two sub-expert sets. CartesianMoE enables group-wise knowledge sharing among experts and helps to build a more complete knowledge sharing system with shared experts equipped.
    \item  We validate the effectiveness of the proposed CartesianMoE with extensive experiments. Experimental results show that it consistently outperforms previous MoE models, and shows better routing robustness.
\end{itemize}

\section{Related Work}
The concept of MoE models was first introduced by~\cite{DBLP:journals/neco/JacobsJNH91}. Then, ~\cite{eigen2013learning} extended the MoE model to multiple layers. Later, ~\cite{DBLP:conf/iclr/ShazeerMMDLHD17} extended that idea to Long Short-Term Memory (LSTM) networks~\cite{DBLP:journals/corr/Graves13}, training an LSTM model with up to 137 billion parameters. With the advent of the Transformer architecture~\cite{DBLP:conf/nips/VaswaniSPUJGKP17,DBLP:conf/naacl/DevlinCLT19}, the Gshard model~\cite{DBLP:conf/iclr/LepikhinLXCFHKS21} applied MoE techniques to Transformers, paving the way for the development of more advanced MoE models like GLaM~\cite{DBLP:conf/icml/DuHDTLXKZYFZFBZ22} and Switch Transformer~\cite{DBLP:journals/jmlr/FedusZS22}.

In early works~\cite{zoph2022st,DBLP:journals/jmlr/FedusZS22,DBLP:conf/icml/DuHDTLXKZYFZFBZ22,DBLP:conf/iclr/LepikhinLXCFHKS21,DBLP:conf/nips/RollerSSW21,DBLP:conf/acl/Dai0MZSCW22}, when extending dense models to MoE models, the MoE layer of the Transformer consists of multiple FFNs that are of the same size as those in the dense models. Recent works~\cite{muennighoff2024olmoe,DBLP:journals/corr/abs-2401-06066,yang-etal-2024-xmoe} show that splitting a fully-sized FFN into several smaller, fine-grained experts facilitates representation specialization. Additionally, shared experts are commonly adopted~\cite{DBLP:journals/corr/abs-2401-06066,DBLP:conf/icml/RajbhandariLYZA22,su2024maskmoe}, to enhance knowledge sharing among experts for performance improvement.

The shared-expert method combines shared knowledge (i.e., the outputs of the shared experts) with specialized knowledge (i.e., the outputs of the routed experts) in an ``addition'' manner. Inspired by collective matrix factorization to learn shared knowledge among data, here we propose an alternative ``multiplication'' manner to share expert knowledge, which demonstrates superiority over previous MoE methods.



\section{Background}

\subsection{Large Language Models}
For simplicity, here we focus on the mainstream \textit{generative LLM} with the Transformer backbone. Given a sequence of $T$ tokens \(\mathbf{x}=({x_{1}, x_{2}, \ldots, x_{T}})\), a generative LLM iteratively produces a probability distribution \(\mathbf{p}\) over the vocabulary for each token, conditioning on its preceding tokens. Usually, the cross-entropy loss function is employed to optimize the predicted probability w.r.t the ground-truth token \(x_t\). And thus in total, the training loss \(\mathcal{L}_{lm}\) for the generative LLM can be expressed as:
\begin{equation}
\begin{aligned}
    \mathcal{L}_{lm} &= - \sum_{t=1}^{T-1}  \log(\mathbf{P}_{x_{t+1}, t}) \\
    \text{s.t.,} \quad \mathbf{P}_{\cdot, t} &= \text{softmax}(W \mathbf{H}^L_{\cdot, t}) \\
    \mathbf{H}^{L} &= \text{Transformer}(x_{1},x_{2}, \ldots, x_{T-1})
\end{aligned}
\end{equation}
Here, $L$ is the number of blocks in the Transformer backbone. \(\mathbf{P}_{\cdot, t}\) and \(\mathbf{H}_{\cdot, t}^L\) represent the \(t\)-th column of the matrices \(\mathbf{P}\) and \(\mathbf{H}^L\), respectively, corresponding to $x_t$. $\mathbf{H}^L = [\mathbf{h}_{1}^L,\mathbf{h}_{2}^L, \ldots, \mathbf{h}_{T-1}^L]$ denotes the hidden states of the last layer, and \(\mathbf{P}_{x_{t+1}, t}\) denotes the predicted probability w.r.t the ground-truth token \(x_{t+1}\) in \(\mathbf{P}_{\cdot, t}\). Here the linear projection layer \(W\) takes \(\mathbf{H}^L_{\cdot, t}\) as input to compute the probability distribution \(\mathbf{P}_{\cdot, t}\) across the vocabulary.

In the Transformers backbone, each layer features a multi-head self-attention (MHA) module and a feed-forward network (FFN), with the FFN typically comprising two fully connected layers. Formally,
\begin{equation}
\begin{aligned}
    \mathbf{\hat{h}}_{t}^{l} &= \text{MHA}([\mathbf{h}_{1}^{l-1}, \mathbf{h}_{2}^{l-1}, \ldots, \mathbf{h}_{t}^{l-1}]) \\ 
    \mathbf{h}_{t}^{l} &= \text{FFN}(\mathbf{\hat{h}}_{t}^{l})
\end{aligned}
\end{equation}
where \(l\) denotes the \(l\)-th block in the Transformer backbone. 

\subsection{Mixture-of-Experts}

MoE methods typically replace the dense model's FFN module with an MoE module composed of multiple FFNs, each being an \textit{expert}. The outputs of these FFNs are combined using a routing function, \(\mathbf{r}(\cdot)\), referred to as the \textit{router}. Formally,
\begin{equation}
\begin{aligned}
\label{full}
\mathbf{h}_{t}^{l} = \sum_{i=1}^{N} \mathbf{r}_{i}(\mathbf{\hat{h}}_{t}^{l}) \cdot \text{FFN}_{i}(\mathbf{\hat{h}}_{t}^{l}) 
\quad  \text{s.t.} \,\, \vert \mathbf{r}(\mathbf{\hat{h}}_t^l)\vert_0 = K
\end{aligned}
\end{equation}
where \(N\) is the number of experts in a single MoE module, \(K\) is the number of activated experts, \(\mathbf{r}_i\) represents the routing outcome for the \(i\)-th expert, and \(\vert \cdot \vert_0\) denotes the \(L_0\)-norm, i.e., the number of non-zero elements. With $K \ll N$, only a small subset of experts is activated. And thus increasing the total number of experts in MoE models does not significantly increase computational time. 

For fine-grained experts~\cite{muennighoff2024olmoe,DBLP:journals/corr/abs-2401-06066,yang-etal-2024-xmoe}, each of the original \(N\) experts is split into \(m\) equal parts, resulting in \(mN\) fine-grained experts in total. In that case, the intermediate size of the fine-grained experts is \( \frac{1}{m} \) of the original full-sized experts. To maintain a constant number of activated parameters, the number of activated experts is usually adjusted to \(mK\) as well. Formally,
\begin{equation}
\begin{aligned}
\label{fine}
\mathbf{h}_{t}^{l} = \sum_{i=1}^{mN} \mathbf{r}_{i}(\mathbf{\hat{h}}_{t}^{l}) \cdot \text{FFN}_{i}(\mathbf{\hat{h}}_{t}^{l}) 
\quad  \text{s.t.} \,\, \vert \mathbf{r}(\mathbf{\hat{h}}_t^l)\vert_0 = mK
\end{aligned}
\end{equation}



\section{Method}

\subsection{Proposed CartesianMoE}
As mentioned above, the current MoE models either rarely share learned knowledge among experts or only apply shared experts to share global knowledge. We propose CartesianMoE, as shown in Fig.~\ref{fig:overall_fig}, to facilitate more thorough expert sharing.

As shown in Figure \ref{fig:subfig3}, the proposed CartesianMoE introduces a \textit{Cartesian Product Layer}, and also employs fine-grained experts in its two MoE sub-layers, denoted as $A$ and $B$. Then CartesianMoE combines the fine-grained sub-experts across the two MoE sub-layers to derive real experts. Formally,
\begin{equation}
\begin{aligned}
A \times B &= \{(a, b) \mid a \in A \text{ and } b \in B\} \\
\text{s.t.} \quad A &= \{\text{FFN}_{1}, \ldots, \text{FFN}_{e}\}, \\
B &= \{\text{FFN}_{e+1}, \ldots, \text{FFN}_{2e}\}.
\end{aligned}
\end{equation}
where $e$ is the number of sub-experts in each MoE sub-layer. To maximize the diversity of $A\times B$, we set \( e = mN/2 \) experts, with $mN$ being the number of all fine-grained sub-experts. Specifically, the computation of the \textit{Cartesian Product Layer} is formulated as follows. 
\begin{flalign}
\label{ce:1}
& \mathbf{\tilde{h}}_{t}^{l} = \sum_{i=1}^{e} \mathbf{r}^{1}_{i}(\mathbf{\hat{h}}_{t}^{l}) \cdot \text{FFN}_{i}(\mathbf{\hat{h}}_{t}^{l}) \\
\label{ce:2}
& \mathbf{\bar{h}}_{t}^{l} = \mathbf{\tilde{h}}_{t}^{l} + \mathbf{\hat{h}}_{t}^{l} \\
\label{ce:3}
& \mathbf{h}_{t}^{l} = \sum_{i=e+1}^{2e} \mathbf{r}^{2}_{i}(\mathbf{\bar{h}}_{t}^{l}) \cdot \text{FFN}_{i}(\mathbf{\bar{h}}_{t}^{l}) \\
\label{ce:4}
& \text{s.t.} \,\, \vert \mathbf{r}^{1}(\mathbf{\hat{h}}_t^l)\vert_0 = \vert \mathbf{r}^{2}(\mathbf{\bar{h}}_t^l)\vert_0 = k 
\end{flalign}
where \( \mathbf{r}^{1}\) and \( \mathbf{r}^{2} \) represent the \textit{routers} corresponding to the 1st and 2nd MoE sub-layers of the \textit{Cartesian Product Layer}, respectively. 
Note that we also add a residual connection between the two MoE sub-layers, to ensure that tokens exceeding the capacity of a sub-expert in the 1st MoE sub-layer can be directly passed to the 2nd MoE sub-layer, i.e., ``token droppable'' in ~\cite{DBLP:journals/jmlr/FedusZS22} to balance optimization among experts. 
In order to maintain a consistent total number of activated parameters as previous fine-grained MoE methods, the number of activated experts per MoE sub-layer, i.e., $k$ in Eq.~\ref{ce:4}, is also reduced by half, 
i.e., $k=mK/2$. We term such a routing process as Cartesian Product routing. Through such a two-layer structural design, the Cartesian Product mechanism is natively implemented, and is supposed to facilitate knowledge sharing among experts. 

Then following Transformer~\cite{DBLP:conf/nips/VaswaniSPUJGKP17}, we add \(\mathbf{\bar{h}}_{t}^{l}\) to \(\mathbf{h}_{t}^{l}\) to serve as the input for the next block with a skip connection. Formally,
\begin{equation}
\begin{aligned}
\label{res}
\mathbf{h}_{t}^{l} \gets \mathbf{h}_{t}^{l} +  \mathbf{\bar{h}}_{t}^{l}
\end{aligned}
\end{equation}


\subsection{Load Balance Loss}
LLMs are typically trained in a distributed manner, which can lead to load imbalances in MoE models~\cite{DBLP:conf/iclr/LepikhinLXCFHKS21, DBLP:journals/jmlr/FedusZS22, DBLP:conf/acl/Dai0MZSCW22}, where a minority of experts handle the majority of tokens and meanwhile the majority of experts remain idle. Such imbalances can adversely affect the training efficiency. 
To address that issue, a load balancing loss is commonly introduced in the training of MoE models. We follow ~\cite{DBLP:journals/corr/abs-2403-07652,DBLP:journals/jmlr/FedusZS22} and employ a balanced loss function by summing the routing losses of both MoE sub-layers within a \textit{Cartesian Product Layer}:
\begin{equation}
\begin{aligned}
\label{eq}
\mathcal{L}_{bal} &= \sum_{i=1}^{e} w^{1}_{i} R^{1}_{i} + \sum_{i=e+1}^{2e} w^{2}_{i} R^{2}_{i} \\
\text{s.t.},\quad  
w^{k}_{i} &= \frac{1}{B} \sum_{j=1}^{B} \mathbb{I} \left\{ \text{argmax}(\mathbf{r}_{\cdot,j}^{k}) = i \right\} \\
R^{k}_{i} &= \frac{1}{B} \sum_{j=1}^{B} \mathbf{r}^{k}_{i,j} \\
\forall &\ k \in \{1, 2\}
\end{aligned}
\end{equation}
where $B$ represents the number of tokens in a mini-batch, $k\in\{1,2\}$ denotes the sub-layer index within a \textit{Cartesian Product Layer}, $\mathbf{r}_{\cdot,j}^{k}$ denotes the routing output probability distribution for the $j$-th token in the 1st ($k=1$) or 2nd ($k=2$) MoE sub-layer, and $\mathbf{r}_{i,j}^{k}$ represents the specific probability value with respect to the $i$-th expert in either the 1st ($k=1$) or 2nd ($k=2$) MoE sub-layer.

Our final loss is a combination of the language model loss and the load-balance loss:
\begin{align}
\label{eq:loss}
    \mathcal{L} = \mathcal{L}_{lm} +  \alpha\mathcal{L}_{bal}
\end{align}
where $\alpha$ is a hyperparameter.

\subsection{Relations to Flattened Fine-grained Experts}
\label{section:relations}
As detailed above, the proposed CartesianMoE leverages two layers of fine-grained sub-experts to build a \textit{Cartesian Product Layer}. Then it would be interesting to see its relations to the flattened fine-grained experts proposed in~\cite{DBLP:journals/corr/abs-2401-06066}. 

Suppose the number of fine-grained experts/sub-experts is $2e$ for both methods, same as before. The output of the MoE module in~\cite{DBLP:journals/corr/abs-2401-06066}, together with that of the residual connection, is formulated as below:
\begin{equation}
\label{ce:5}
\begin{aligned}
 \mathbf{h}_{t}^{l} &= \mathbf{\hat{h}}^l_t + \sum_{i=1}^{e} \mathbf{r}^{1}_{i}(\mathbf{\hat{h}}^l_t) \cdot \text{FFN}_{i}(\mathbf{\hat{h}}^l_t) \\
 & + \sum_{i=e+1}^{2e} \mathbf{r}^{1}_{i}(\mathbf{\hat{h}}^l_t) \cdot \text{FFN}_{i}(\mathbf{\hat{h}}^l_t) 
\end{aligned}
\end{equation}
where $\mathbf{r}^{1}$ denotes the single \textit{router} in the MoE module. As for the proposed CartesianMoE, the output of the \textit{Cartesian Product Layer}, can be derived as below, via integrating Eq.~\ref{ce:1}, Eq.~\ref{ce:2} and Eq.~\ref{ce:3} into Eq.~\ref{res}.  
\begin{equation}
\label{ce:6}
\begin{aligned}
 \mathbf{h}_{t}^{l} &= \mathbf{\hat{h}}^l_t + \sum_{i=1}^{e} \mathbf{r}^{1}_{i}(\mathbf{\hat{h}}^l_t) \cdot \text{FFN}_{i}(\mathbf{\hat{h}}^l_t) \\
 & + \sum_{i=e+1}^{2e} \mathbf{r}^{2}_{i}(\mathbf{\bar{h}}^l_t) \cdot \text{FFN}_{i}(\mathbf{\bar{h}}^l_t) 
\end{aligned}
\end{equation}

Comparing Eq.~\ref{ce:5} and Eq.~\ref{ce:6}, it can be seen that the proposed CartesianMoE and the flattened fine-grained experts mainly differ at the 3rd parts of both equations, i.e., $\sum_{i=e+1}^{2e} \mathbf{r}^{1}_{i}(\mathbf{\hat{h}}^l_t) \cdot \text{FFN}_{i}(\mathbf{\hat{h}}^l_t)$ versus $\sum_{i=e+1}^{2e} \mathbf{r}^{2}_{i}(\mathbf{\bar{h}}^l_t) \cdot \text{FFN}_{i}(\mathbf{\bar{h}}^l_t)$. Specifically, instead of sharing the same \textit{router} $\mathbf{r}^1$ and the same input $\mathbf{\hat{h}}^l_l$ as the 2nd part, CartesianMoE leverages a separate \textit{router} $\mathbf{r}^2$ and the output of the 1st sub-layer as input, i.e., $\mathbf{\bar{h}}_t^l$. Given $\mathbf{\bar{h}}_{t}^{l} = \mathbf{\tilde{h}}_{t}^{l} + \mathbf{\hat{h}}_{t}^{l}$ in Eq.~\ref{ce:2}, CartesianMoE can probably enjoy deeper representations of the input than the flattened counterpart, and the separate \textit{router} offers more flexibility. Both can help CartesianMoE to achieve performance enhancement, as demonstrated in our experiments.

\section{Experiments}
\subsection{Pre-training Dataset}
Following previous works~\cite{DBLP:conf/nips/Xie0DDLLLL0Y23,su-etal-2024-mile}, we use the Pile dataset~\cite{DBLP:journals/corr/abs-2101-00027} as our pre-training data. The Pile is a large-scale, publicly available corpus comprising 22 domains and over 825 GB of English text. For tokenization, we utilize the widely adopted LLaMA tokenizer with a vocabulary size of 32k. We compute the sampling rate for each domain based on the number of tokens after tokenization, following the methodology described in ~\cite{DBLP:conf/nips/Xie0DDLLLL0Y23, su-etal-2024-mile}. Due to our limited computational resources, unless otherwise specified, the models are pre-trained using 100B tokens, following ~\cite{DBLP:journals/corr/abs-2401-06066, su2024maskmoe, DBLP:conf/nips/Xie0DDLLLL0Y23, su-etal-2024-mile, DBLP:journals/corr/abs-2403-07652, xiong2024temporal, lian2024scaffold}.


\begin{table*}[!t]
\centering
\scalebox{0.7}{
\begin{tabular}
{llccccc}
\toprule
Model & Configuration & SE & FGE & Params & Activated Params & Pile PPL ($\downarrow$) \\
\midrule
Base Model & d=768, D=3072 & N/A & N/A & 162M & 162M & 8.55  \\
Large Model & d=1024, D=4096 & N/A & N/A & 468B & 468M & 6.95  \\
\midrule
\textit{MoE-Base} \\
\midrule 
SMoE-Share & d=768, D=3072, top$K$=2 & True & False & 842M & 247M & 7.37  \\
SMoE-Top3 & d=768, D=3264, top$K$=3 & False & False & 842M & 258M & 7.40  \\
Hash Layer & d=768, D=3072, top$K$=2 & True & False & 842M & 247M & 7.47  \\
Fine-grained Routing& d=768, D=1536, top$K$=4, & True & True & 842M & 247M & 7.33  \\
TopP Routing & d=768, D=3072, top$P$=0.4 & True & False & 842M & 247M & 7.41 \\
CartesianMoE & d=768, D=1526, top$K$=(2+2) & True & True & 842M & 247M & \textbf{7.19} \\
\midrule
\textit{MoE-Large} \\
\midrule
SMoE-Share & d=1024, D=4096, top$K$=2 & True & False & 2.88B & 770M & 6.13  \\
SMoE-Top3 & d=1024, D=4352, top$K$=3 & False & False & 2.88B & 808M & 6.18  \\
Hash Layer & d=1024, D=4096, top$K$=2 & True & False & 2.88B & 770M & 6.28  \\
Fine-grained Routing& d=1024, D=2048, top$K$=4 & True & True & 2.88B & 770M & 6.16  \\
TopP Routing& d=1024, D=4096, top$P$=0.4 & True & False & 2.88B & 770M & 6.14 \\
CartesianMoE & d=1024, D=2048, top$K$=(2+2) & True & True & 2.88B & 770M & \textbf{6.08} \\
\bottomrule
\end{tabular}
}
\caption{ Perplexity (PPL) results of language modeling. The best score is marked in \textbf{bold}. SE indicates whether to use shared experts, FGE indicates whether to use fine-grained experts, \text{d} represents the hidden state dimensionality, \text{D} represents the intermediate size of each FFN, and top$K$ refers to the number of experts activated for each token. For CartesianMoE, top$K$=(2+2) means that each of the two sub-layers activates two sub-experts. For TopP Routing, top$P$ is the threshold that controls how many experts should be activated to reach it. }
\label{PPL_results}
\end{table*}

\subsection{Experimental Setup}
Following ~\cite{yang-etal-2024-xmoe,DBLP:journals/corr/abs-2403-07652}, we implement the LLaMA architecture for the \textit{LARGE} models with 24 Transformer blocks and a hidden state dimensionality of 1024, and for the \textit{BASE} models with 12 Transformer blocks and a hidden-state dimensionality of 768. We employ the AdamW ~\cite{DBLP:conf/iclr/LoshchilovH19} optimizer for all models with a cosine learning rate decay schedule. For the dense models, following ~\cite{DBLP:journals/corr/abs-2302-13971,su-etal-2024-mile}, we set the learning rate as $3e^{-4}$. For the MoE models, following ~\cite{DBLP:conf/icml/LewisBDGZ21,su-etal-2024-mile,DBLP:journals/corr/abs-2302-13971}, we reduce the learning rate to $1.5e^{-4}$ to ensure model convergence. 
By default, we set our maximum sequence length to 1024.

Following ~\cite{yang-etal-2024-xmoe,DBLP:journals/corr/abs-2403-07652,su2024maskmoe}, we conduct experiments on two different MoE model settings: \textit{MoE-Base} and \textit{MoE-Large}. The specific size configurations are shown in Table~\ref{PPL_results}. 
We follow Gshard~\cite{DBLP:conf/iclr/LepikhinLXCFHKS21}, and replace the FFN layer with an MoE layer for every other Transformer block, resulting in a total of $12$ MoE layers for \textit{MoE-Large} and $6$ MoE layers for \textit{MoE-Base} in this setting. For the hyperparameter \(\alpha\) w.r.t the load balanced loss (Eq.~\ref{eq:loss}), we set it to $0.01$. The expert capacity factor of tokens is set as $1$ during training. Moreover, we adopt a dropless setup, ensuring that every token is retained during evaluation.

In the CartesianMoE, each \textit{Cartesian Product Layer} contains 32 fine-grained sub-experts, with each sub-expert having a half-sized FFN. We assign 16 fine-grained sub-experts to each of the two MoE sub-layers, and use top-2 routing for each. In addition, each MoE sub-layer has a fixed-activated shared expert, so as to form a ``global shared knowledge + group-wise shared knowledge + expert-specific knowledge'' system mentioned before. We compare the proposed CartesianMoE with 6 remarkable baselines~\cite{DBLP:journals/corr/abs-2302-13971,DBLP:journals/jmlr/FedusZS22,DBLP:conf/nips/RollerSSW21,DBLP:journals/corr/abs-2403-07652,DBLP:journals/corr/abs-2401-06066} in our experiments. 
The respective parameter settings for each compared model are provided in Appendix ~\ref{subsec:compared_models}. Considering that shared experts are commonly included in MoE models~\cite{DBLP:journals/corr/abs-2401-06066,DBLP:journals/corr/abs-2405-04434,zhao2024hypermoe,DBLP:conf/icml/RajbhandariLYZA22,su2024maskmoe}, all compared baselines include shared experts to gain further performance improvement unless otherwise noted.

\subsection{Main Results}
We first present the model's perplexity (PPL) on the Pile validation set. Then, following \cite{DBLP:journals/corr/abs-2302-13971, DBLP:conf/nips/BrownMRSKDNSSAA20, su-etal-2024-mile, DBLP:journals/corr/abs-2401-06066}, we evaluate the model performance on various downstream benchmarks, including zero-shot tests for HellaSwag \cite{DBLP:conf/acl/ZellersHBFC19}, LAMBADA \cite{DBLP:conf/acl/PapernoKLPBPBBF16}, PIQA \cite{DBLP:conf/aaai/BiskZLGC20},  StoryCloze \cite{DBLP:journals/corr/MostafazadehCHP16}, and Winogrande(Wino) \cite{DBLP:conf/aaai/SakaguchiBBC20}, in terms of \textit{accuracy}. In addition, following \cite{DBLP:journals/corr/abs-2302-13971, su-etal-2024-mile}, we conduct 5-shot evaluations on TriviaQA \cite{DBLP:conf/acl/JoshiCWZ17}, WebQuestions (WebQs) \cite{DBLP:conf/emnlp/BerantCFL13}, and Natural Questions (NaturalQs) \cite{DBLP:journals/tacl/KwiatkowskiPRCP19} using the \textit{exact match} metric. 

\begin{table*}[!t]
\centering
\scalebox{0.65}{
\begin{tabular}{lp{1.40cm}<{\centering}p{1.40cm}<{\centering}p{1.40cm}<{\centering}p{1.40cm}<{\centering}p{1.40cm}<{\centering}p{1.40cm}<{\centering}p{1.40cm}<{\centering}p{1.40cm}<{\centering}p{1.40cm}<{\centering}}
\toprule
 Model & Hellaswag & LAMBADA & PIQA & StoryCloze & Wino & TriviaQA & WebQs & NatrualQs \\
\midrule
Base Model   & 32.54 & 39.22 & 62.52  & 58.52 & 50.75 & 2.99 & 2.22  & 0.72 \\
Large Model   & 40.73 & 52.55 & 67.62 & 63.55 &53.75 & 7.44 & 4.97 & 1.78  \\
\midrule
\textit{MoE-Base} \\
\midrule
SMoE-Share & 37.87 & 47.35 & \textbf{\underline{66.76}}  & 61.52 & 51.14 & 6.11 & \textbf{\underline{4.18}} & 1.39  \\
SMoE-Top3 & 37.43 & 46.42 & 66.38  & 61.25 & 51.14 & 5.82  & 4.04 & 1.16 \\
Hash Layer  & 33.39 &43.08 & 61.59 & 57.88 & 50.20 & 2.85 & 1.57 & 0.61 \\
Fine-grained Routing & 37.50 & 45.10 & 66.49  & 61.20 & 51.22 & 6.12  & 3.79 & 1.30  \\
TopP Routing & 35.10 & 45.02 & 63.76 & 58.10 & 51.46 & 4.12  & 2.41 & 0.55  \\
CartesianMoE & \textbf{\underline{38.17}} & \textbf{\underline{48.17}} & 65.83 & \textbf{\underline{62.53}} & \textbf{\underline{51.62}} & \textbf{\underline{7.23}}  & \textbf{\underline{4.18}} & \textbf{\underline{1.58}} \\
\midrule
\textit{MoE-Large} \\
\midrule
SMoE-Share & 48.25 & 58.80 & \textbf{\underline{70.51}} & 66.49 & 55.56 & 14.91 & 9.06 & 3.99 \\
SMoE-Top3 & 47.59 & 57.60 & 70.40 & 65.95 & 53.75 & 14.46 & 7.92 & 4.24  \\
Hash Layer  & 42.80 & 56.45 & 66.32  & 64.08 & 52.33 & 6.66 & 4.33 & 1.86 \\
Fine-grained Routing & 48.10 & 58.32 & 69.86  & 64.89 & 55.80 & 14.40  & 8.42 & 3.77 \\
TopP Routing & 44.31 & \textbf{\underline{59.85}} & 67.46 & 63.92 & 54.54 & 11.02  & 6.15 & 2.41 \\
CartesianMoE & \textbf{\underline{49.14}} & 59.17 & 70.24 & \textbf{\underline{67.02}} & \textbf{\underline{56.51}} & \textbf{\underline{15.09}}  & \textbf{\underline{9.15}} & \textbf{\underline{4.85}} \\
\bottomrule
\end{tabular}
}
\caption{Performances of language models on downstream tasks. The best score is marked in \textbf{\underline{bold}}.}
\label{expert_results}
\end{table*}

\subsubsection{Perplexity Results}
Table \ref{PPL_results} shows the perplexity (PPL) of language modeling on the Pile validation set. With the \textit{same number of activated parameters}, the MoE models (\textit{MoE-Base/MoE-Large}) consistently outperform the dense models (\textit{Base/Large Model}) with significantly reduced PPL. 
Furthermore, CartesianMoE exhibits a substantial performance improvement over other models, in both \textit{MoE-Base} and \textit{MoE-Large} settings. 
The result presents the superiority of CartesianMoE, which is equipped with complete ``global shared knowledge + group-wise shared knowledge + expert-specific knowledge''.

Note that \textit{Fine-grained Routing} with flattened fine-grained experts exhibits inconsistent improvements across different model sizes. In the \textit{MoE-Base} setting, it significantly outperforms \textit{SMoE-Share}, but in the \textit{MoE-Large} setting, it performs slightly worse than \textit{SMoE-Share}. In contrast, CartesianMoE demonstrates consistent performance improvements across different settings, highlighting its consistent superiority.

\subsubsection{Benchmark Results}
As shown in Table \ref{expert_results}, we present the model's performance on downstream tasks. We can also observe that the MoE models achieve performance improvements over the dense counterpart in those benchmark tasks. More importantly, the proposed CartesianMoE stands out among all MoE models, in both \textit{MoE-Base} or \textit{MoE-Large} settings.
Specifically, compared to other MoE models, CartesianMoE yields the best performance for 7 of 8 benchmarks in the \textit{MoE-Base} setting and 6 of 8 benchmarks in the \textit{MoE-Large} setting. 

Particularly, against \textit{Fine-grained Routing} with flattened fine-grained experts, CartesianMoE excels in 7 of 8 benchmarks with the \textit{MoE-Base} setting and also excels in all benchmarks with the \textit{MoE-Large} setting. That further verifies our analysis above that CartesianMoE can enjoy deeper representations of input and more flexible routing than the flattened counterpart. It also demonstrates the effectiveness of introducing the \textit{Cartesian Product Layer} for group-wise knowledge sharing. 




\section{Analyses}


\subsection{Impact of Fixed-Activated Shared Expert}
Under the \textit{MoE-Large} setting, we remove the fixed-activated shared experts from CartesianMoE to investigate its impact on the model performance. 

As shown in Table ~\ref{shared_expert}, after removing the fixed-activated shared experts (i.e., \textit{w/o Shared Expert}), CartesianMoE yields slightly better performance than \textit{Fine-grained Routing} equipped with shared experts. The result well reflects the effectiveness of group-wise knowledge sharing among experts proposed by CartesianMoE, which is equally important as global knowledge sharing introduced by shared experts. Moreover, when CartesianMoE is equipped with shared experts as by default, its performance is substantially enhanced, which further demonstrates the effectiveness of forming a ``global shared knowledge + group-wise shared knowledge + expert-specific knowledge'' system, as proposed by CartesianMoE. 

\subsection{Analysis on Expert Routing Robustness}
To analyze the expert routing robustness of different MoE models, we disable the top-$1$ routed expert and then evaluate the PPL variance brought by such a routing change on the Pile validation set. Specifically, for each token, we mask the expert with the highest routing probability and then select the top $K$ experts from the remaining ones. 
Since each \textit{Cartesian Product Layer} in CartesianMoE has two MoE sub-layers, we randomly select one sub-layer each time and mask the corresponding top-1 expert. 

As shown in Table~\ref{ppl_disable_top1}, even with the top-1 routed expert disabled, CartesianMoE still yields the lowest PPL, and enjoys a much smaller PPL variance, compared to other MoE methods. That well indicates the superior routing robustness of CartesianMoE. And we attribute it to the more thorough knowledge sharing among experts in CartesianMoE, which includes both global and group-wise knowledge sharing. 
\begin{table}[!t]
\centering
\scalebox{0.7}{
\begin{tabular}
{lc}
\toprule
 Model  & PPL \\
\midrule
SMoE-Top3 & 6.18 \\
Fine-grained Routing  & 6.16 \\
CartesianMoE  & \textbf{6.08} \\
\quad \textit{w/o Shared Expert}  & 6.15 \\
\bottomrule
\end{tabular}
}
\caption{Impact of the fixed-activated shared expert.}
\label{shared_expert}
\end{table}

\begin{table}[!t]
\centering
\scalebox{0.7}{
\begin{tabular}{lcc}
\toprule  
 & PPL & PPL(disable top-$1$) \\
\midrule 
    SMoE-Share & 6.13  & 7568.85 \\
    SMoE-Top$3$ & 6.18  & 847.46 \\
    Fine-grained Routing & 6.16 & 3095.44 \\
    TopP Routing & 6.14  & 84.42 \\
    CartesianMoE & \textbf{6.08}  & \textbf{27.72} \\
\midrule
\end{tabular}
}
\caption{PPL on the Pile validation set, with the top-1 routed expert disabled. The baseline \textit{Hash Layer} is excluded here, as the experts for each input in it are fixedly assigned.}
\label{ppl_disable_top1}
\end{table}



\subsection{Training with More Tokens}
The previous experiments are conducted using 100B tokens. 
To investigate whether the superiority of the proposed CartesianMoE can be maintained after training with more tokens, here we continue to train CartesianMoE and the most competitive baseline \textit{Fine-grained Routing}~\cite{DBLP:journals/corr/abs-2401-06066} until 400B tokens, and compare their performance in the \textit{MoE-Large} setting.  

As shown in the left part of Table ~\ref{different_model_size}, on the Pile validation set, the PPL of \textit{Fine-grained Routing} converged to 5.78, while that of CartesianMoE further decreases to 5.69. And on downstream tasks, CartesianMoE also outperforms \textit{Fine-grained Routing} in 6 out of 8 benchmarks. The full changing curves for PPL and benchmark performance of both MoE models are provided in Figure \ref{fig:PPL_schdule} and Figure \ref{fig:down_schedule} in the Appendix, respectively. It can be seen that even trained on more tokens, CartesianMoE consistently maintains superior performance, well demonstrating its effectiveness. 

\begin{table}[!t]
\centering
\scalebox{0.65}{
\begin{tabular}{lcc|cc}
\toprule  
 & \multicolumn{2}{c|}{MoE-Large} & \multicolumn{2}{c}{7.25B Params} \\
 & CartesianMoE & Fine-grain & CartesianMoE & Fine-grain \\
\midrule 
    Hellaswag     & \textbf{54.28}  & 52.10  & \textbf{56.96}  & 55.64 \\
    LAMBADA       & \textbf{63.05}  & 61.55  & \textbf{63.50}  & 62.83 \\
    PIQA          & \textbf{71.71}  & 71.21  & \textbf{73.99}  & 71.87 \\
    StoryCloze    & 67.93 & \textbf{67.97}  & \textbf{69.37} & 69.11 \\
    Wino          & \textbf{56.74}  & 55.88  & \textbf{58.17}  & 58.16 \\
    TriviaQA      & 20.03  & \textbf{21.40}  & \textbf{24.87}  & 22.82 \\
    WebQs         & \textbf{10.58}  & 9.01   & \textbf{11.96}  & 11.32 \\
    NatrualQs     & \textbf{5.87}   & 5.54   & \textbf{6.57}   & 5.54 \\
\midrule
    PPL($\downarrow$) & \textbf{5.69}  & 5.78  & \textbf{4.92}  & 4.99 \\
\midrule
\end{tabular}
}
\caption{The performance comparison after training 400B tokens with different model sizes, with \textit{Fine-grain} being short for the baseline \textit{Fine-grained Routing}. The best score in each setting is marked in \textbf{bold}.}
\label{different_model_size}
\end{table}

\subsection{Scaling Up the Model Size}
To investigate the performance of the proposed CartesianMoE with a larger model size, we follow the setting of~\cite{muennighoff2024olmoe} to train CartesianMoE and the most competitive baseline \textit{Fine-grained Routing}, with 7.25B parameters and 1.61B activated parameters. The specific parameter settings are provided in Appendix ~\ref{subsec:Configuration}. 

As shown in the right part of Table~\ref{different_model_size}, on the Pile validation set, the PPL of \textit{Fine-grained Routing} converged to 4.99, while that of CartesianMoE decreases to 4.92. And on all downstream tasks, CartesianMoE outperforms \textit{Fine-grained Routing}. The full changing curves for PPL and downstream tasks of both MoE models are also provided in Figure ~\ref{fig:PPL_schdule_7b} and Figure ~\ref{fig:down_schedule_7B} in the Appendix. The experimental results further demonstrate the superiority and scalability of CartesianMoE.

\subsection{Training in Different Expert Granularities}

The experiments above use half-sized FFNs as fine-grained experts in CartesianMoE. It would be interesting to see whether CartesianMoE can maintain its superiority with more finer-grained experts. Suppose we have $N$ full-sized experts. As mentioned before, to keep the numbers of total parameters and activated parameters unchanged, we equally split each full-sized expert into $m$ fine-grained experts via splitting its FFN intermediate size into $m$ equal parts, with $m$ being the splitting factor, and the number of activated fine-grained experts would also be scaled up by $m$. It can be seen $m=1$ for full-sized experts, and experiments above use $m=2$ for CartesianMoE. Here we further conduct experiments with $m=4$, for both CartesianMoE and the most competitive baseline \textit{Fine-grained Routing}, to further validate CartesianMoE. 

As is seen in Table~\ref{router_top_k}, in both $m=2$ and $m=4$ settings, CartesianMoE consistently outperforms \textit{Fine-grained Routing} in terms of PPL on the Pile validation set, which further demonstrates its superiority and robustness across different expert granularities. We also find that increasing $m$ may not lead to better performance, as over-fine-grained experts can encounter underfitting. 

\begin{table}[!t]
\centering
\scalebox{0.7}{
\begin{tabular}
{llllcc}
\toprule
D & $K$ & $m$ & Mode  & PPL \\
\midrule
3072 & 2 & 1 & Full-sized & 7.37 \\
1536 & 4 & 2 & Fine-grained Routing  & 7.33 \\
768 & 8 &4 & Fine-grained Routing & 7.34 \\
1536 & 2+2 & 2 & CartesianMoE & \textbf{7.19} \\
768 & 4+4 & 4 & CartesianMoE & 7.26 \\
\bottomrule
\end{tabular}
}
\caption{PPL on the Pile validation set, with different expert granularity. $D$ indicates the FFN intermediate size, $K$ denotes the number of activated experts, and $m$ denotes the splitting factor.}
\label{router_top_k}
\end{table}

\section{Conclusions}
Inspired by collective matrix factorization to capture shared knowledge within data, we introduce CartesianMoE, a ``multiplication''-manner knowledge sharing method among experts in MoE models. CartesianMoE categorizes fine-grained sub-experts into two distinct sets, and uses their Cartesian product to build experts that facilitate group-wise knowledge sharing. Equipped with shared experts as previous works, CartesianMoE builds a more thorough knowledge sharing system among experts, i.e., ``global shared knowledge + group-wise shared knowledge + expert-specific knowledge''. 
Extensive experiments well demonstrate that CartesianMoE outperforms previous MoE models across various settings, in terms of language modeling perplexity and downstream task performance. It also presents much better routing robustness due to enhanced knowledge sharing.

\section{Limitations}
We only perform Cartesian product computations between two MoE sub-layers. In fact, the Cartesian product can be extended to more than two sub-layers. However, ~\cite{DBLP:journals/corr/abs-2407-21783} has shown that increasing the number of model sub-layers requires a corresponding increase in hidden state dimensionality to ensure training effectiveness. 
And thus we leave the exploration of extending to more MoE sub-layers for future work.


\bibliography{anthology,custom}
\bibliographystyle{acl_natbib}
\newpage
\section{Appendix}
\subsection{Compared Models}
\label{subsec:compared_models}
The model settings we compare are as follows. 
For MoE models, following ~\cite{DBLP:conf/iclr/LepikhinLXCFHKS21,yang-etal-2024-xmoe}, unless otherwise specified, each layer of the MoE has 16 experts, with the top-2 experts activated. 
\begin{itemize}[leftmargin=*]
    \item \textbf{Dense} represents a standard Transformer language model.
    \item \textbf{SMoE-Share} denotes an MoE model similar to ~\cite{DBLP:conf/iclr/LepikhinLXCFHKS21,DBLP:journals/jmlr/FedusZS22}, without fine-grained splitting of experts. Additionally, each MoE layer in \textit{SMoE-Share} includes 1 shared expert.
    \item \textbf{SMoE-Top3} denotes an MoE model with top-3 routing and no shared experts. To maintain the total number of parameters after removing the shared expert, \textit{SMoE-Top3} increases the intermediate dimensionality of each expert's FFN, which results in slightly more activated parameters compared to other models and acts as a stronger baseline for comparison.
    \item \textbf{Hash Layer}~\cite{DBLP:conf/nips/RollerSSW21} signifies a method without \textit{router} parameters, where each token is fixedly assigned to two experts using a random hash. The model also has a shared expert for fair comparison with the other models. 
    \item  \textbf{Fine-grained Routing} denotes an MoE model that employs a \textit{Fine-grained Routing} strategy~\cite{DBLP:journals/corr/abs-2401-06066}. For both routing and shared experts, we split the fully-sized FFNs into 2 half-sized FFNs, resulting in 32  fine-grained experts per MoE layer. To maintain the total number of activated parameters consistent, the \textit{Fine-grained Routing} strategy uses top-$4$ routing and includes 2 fixed-activated shared experts for each MoE layer.
    \item  \textbf{TopP Routing}~\cite{DBLP:journals/corr/abs-2403-07652} is a routing strategy that dynamically adjusts the number of activated experts based on the difficulty of tokens. It selects the top experts until their cumulative confidence exceeds the pre-set confidence threshold top$P$. Following ~\cite{DBLP:journals/corr/abs-2403-07652}, we set top$P$ as $0.4$. Similarly, each MoE layer includes one shared expert to enable fair comparison with other models.
\end{itemize}

\newpage

\subsection{Training Configuration}
\label{subsec:Configuration}
\begin{table}[h!]
\centering
\scalebox{0.7}{
\begin{tabular}{lcc}
\hline
 & \textbf{CartesianMoE} & \textbf{Fine-grained Routing} \\ \hline
\textbf{Hidden Size} & 2,048 & 2,048 \\ 
\textbf{Activation} & SwiGLU & SwiGLU \\ 
\textbf{Intermediate Size} & 2,048 & 2,048 \\ 
\textbf{Attn heads} & 16 & 16 \\ 
\textbf{Num layers} & 16 & 16 \\ 
\textbf{Layer norm type} & RMSNorm & RMSNorm \\ 
\textbf{Pos emb.} & RoPE & RoPE \\ 
\textbf{MoE layers} & Every & Every \\  
\textbf{Shared Experts} & True & True \\ 
\textbf{Fine-grained Experts} & True & True \\ 
\textbf{Max seq len} & 4096 & 4096 \\ 
\textbf{MoE sub-layers} & 2 & N/A \\ 
\textbf{\# Experts} & 32 & 32 \\ 
\textbf{\# Activated Expert} & Top$K$=(2+2) & Top$K$=4 \\
\textbf{\# Params} & 7.25B & 7.25B \\ 
\textbf{\# Activated Params} & 1.61B & 1.61B
\\ \hline
\end{tabular}
}
\caption{Configurations of CartesianMoE and \textit{Fine-grained Routing} with 7.25B parameters.}
\end{table}
\newpage

\begin{figure}[h]
    \centering
    \includegraphics[width=0.5\textwidth]{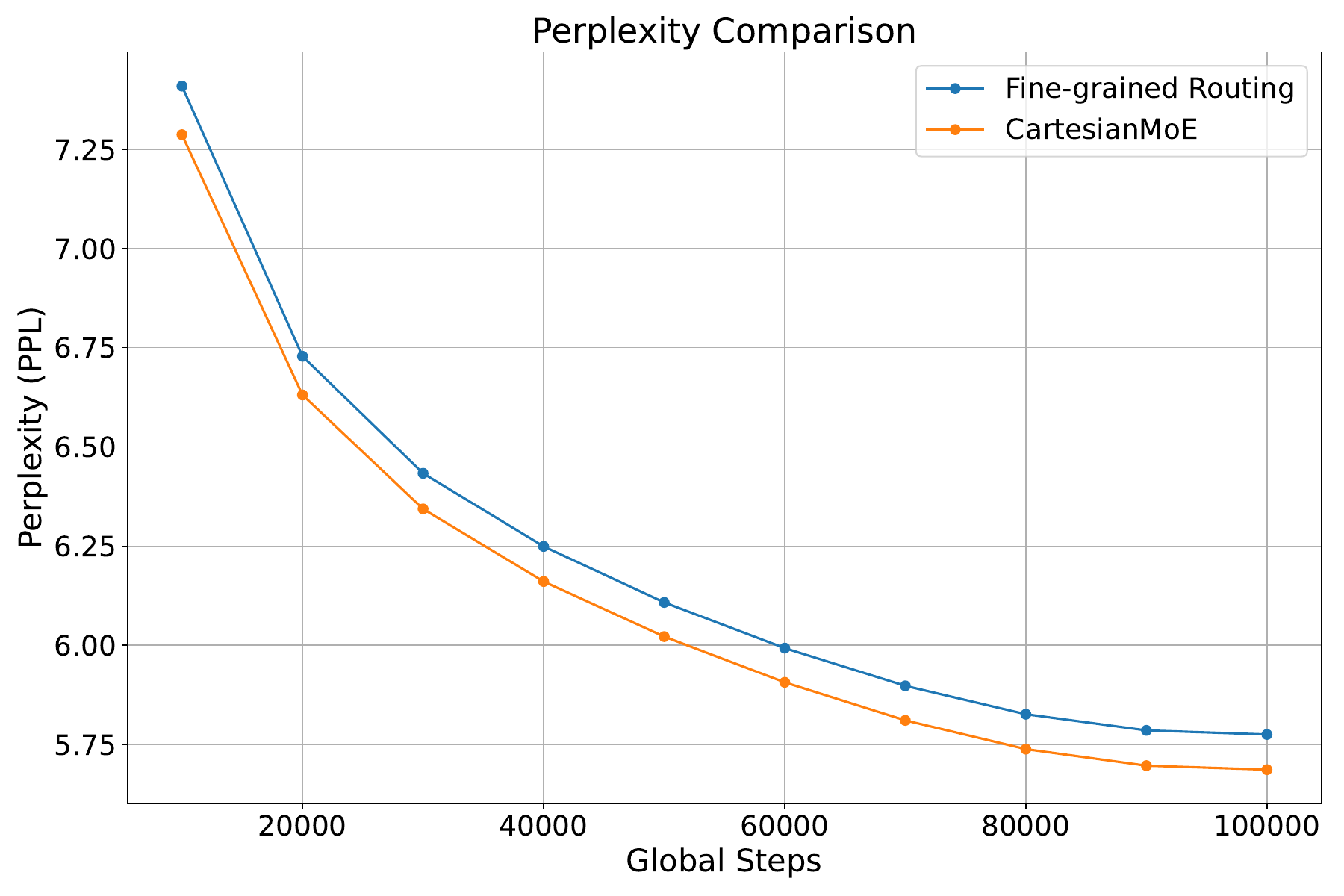}
    \caption{PPL changing curves during language model training with 400B tokens for CartesianMoE and \textit{Fine-grained Routing} in \textit{MoE-Large} setting.}
    \label{fig:PPL_schdule}
\end{figure}

\begin{figure}[h]
    \centering
    \includegraphics[width=0.5\textwidth]{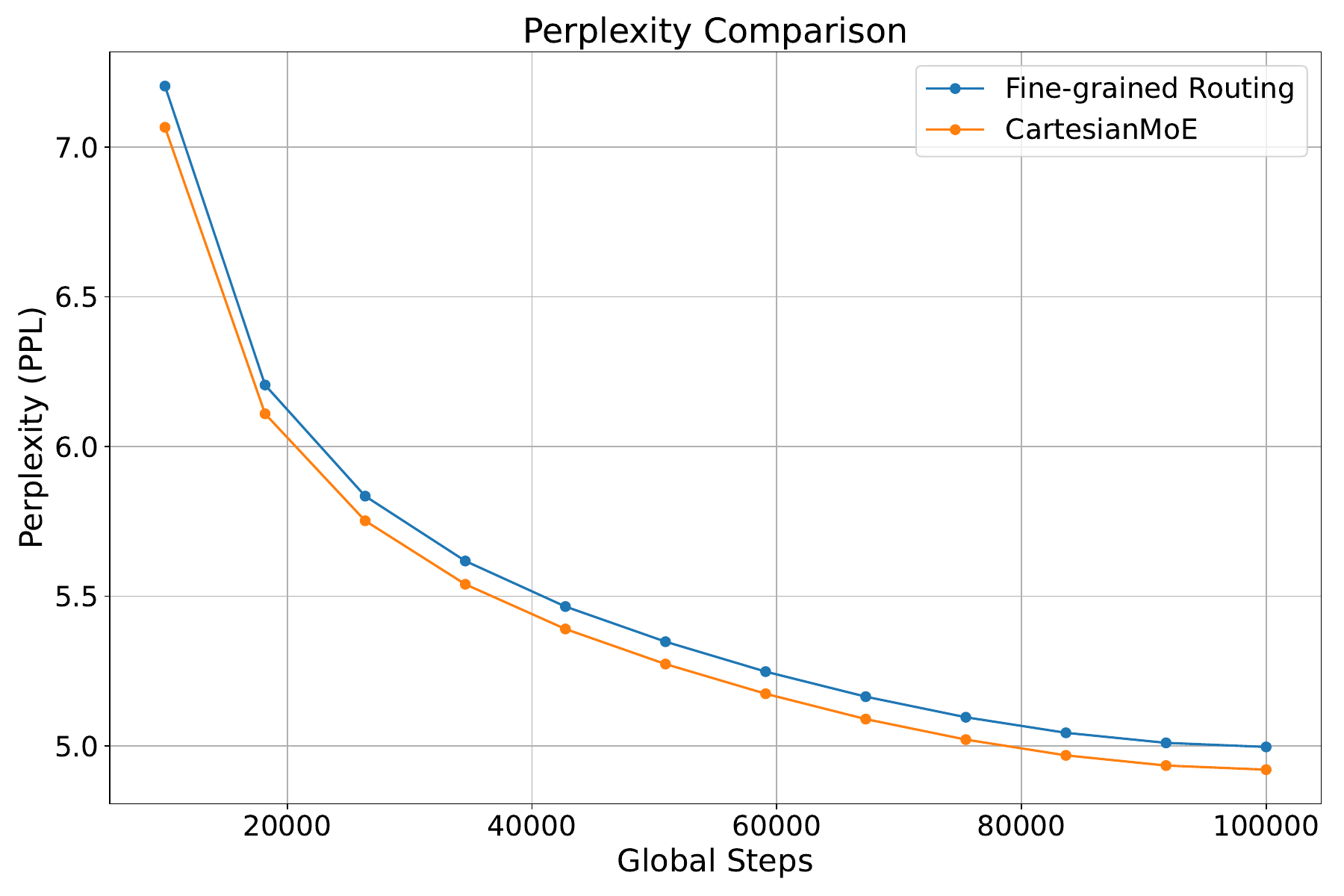}
    \caption{PPL changing curves during language model training with 400B tokens for CartesianMoE and \textit{Fine-grained Routing} with 7.25B parameters.}
    \label{fig:PPL_schdule_7b}
\end{figure}

\begin{figure*}
    \centering
    \includegraphics[width=1\textwidth]{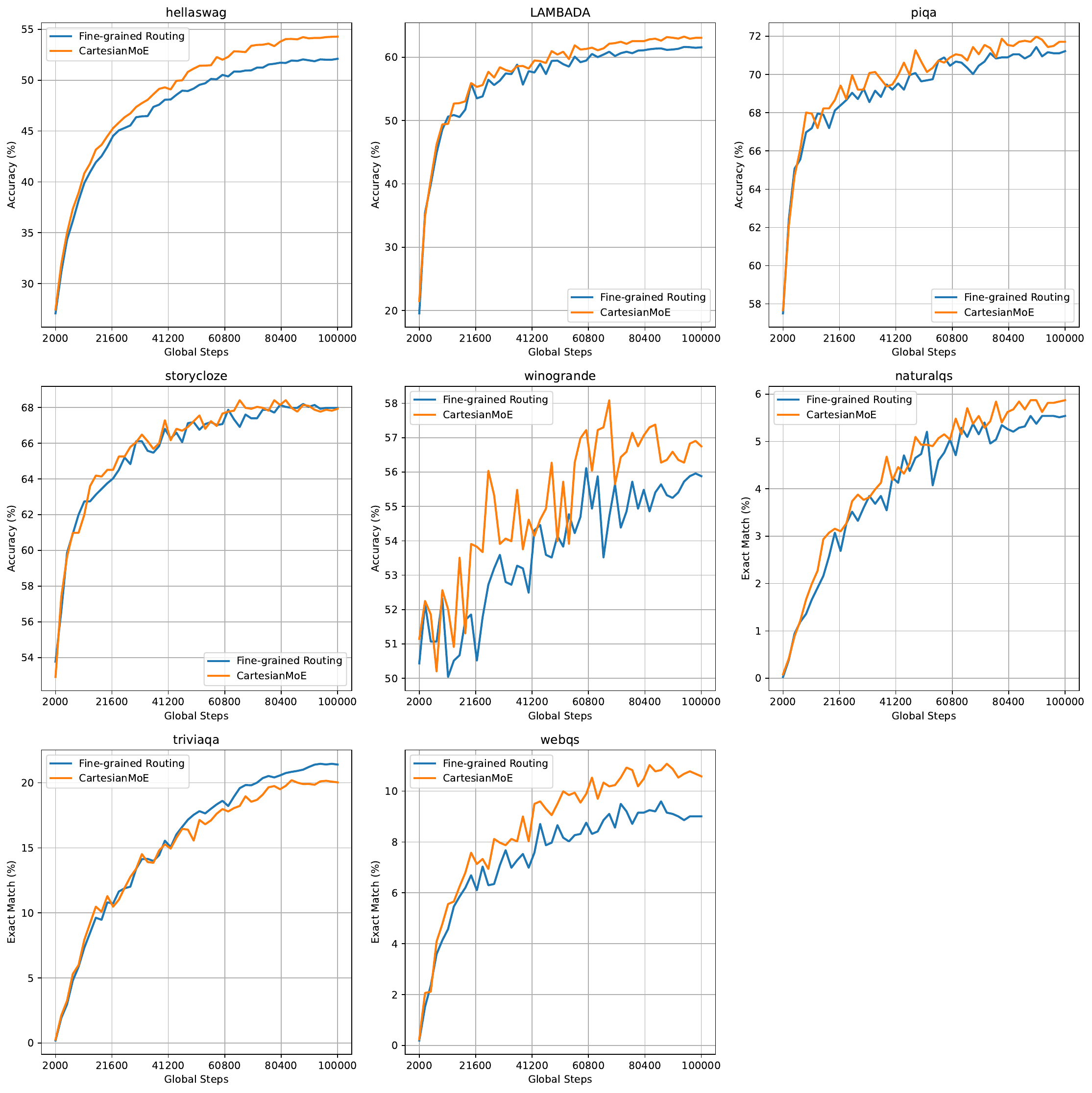}
    \captionof{figure}{Changing curves of downstream task performance during language model training with 400B tokens for CartesianMoE and \textit{Fine-grained Routing} in \textit{MoE-Large} setting.}
    \label{fig:down_schedule}
\end{figure*}

\begin{figure*}
    \centering
    \includegraphics[width=1\textwidth]{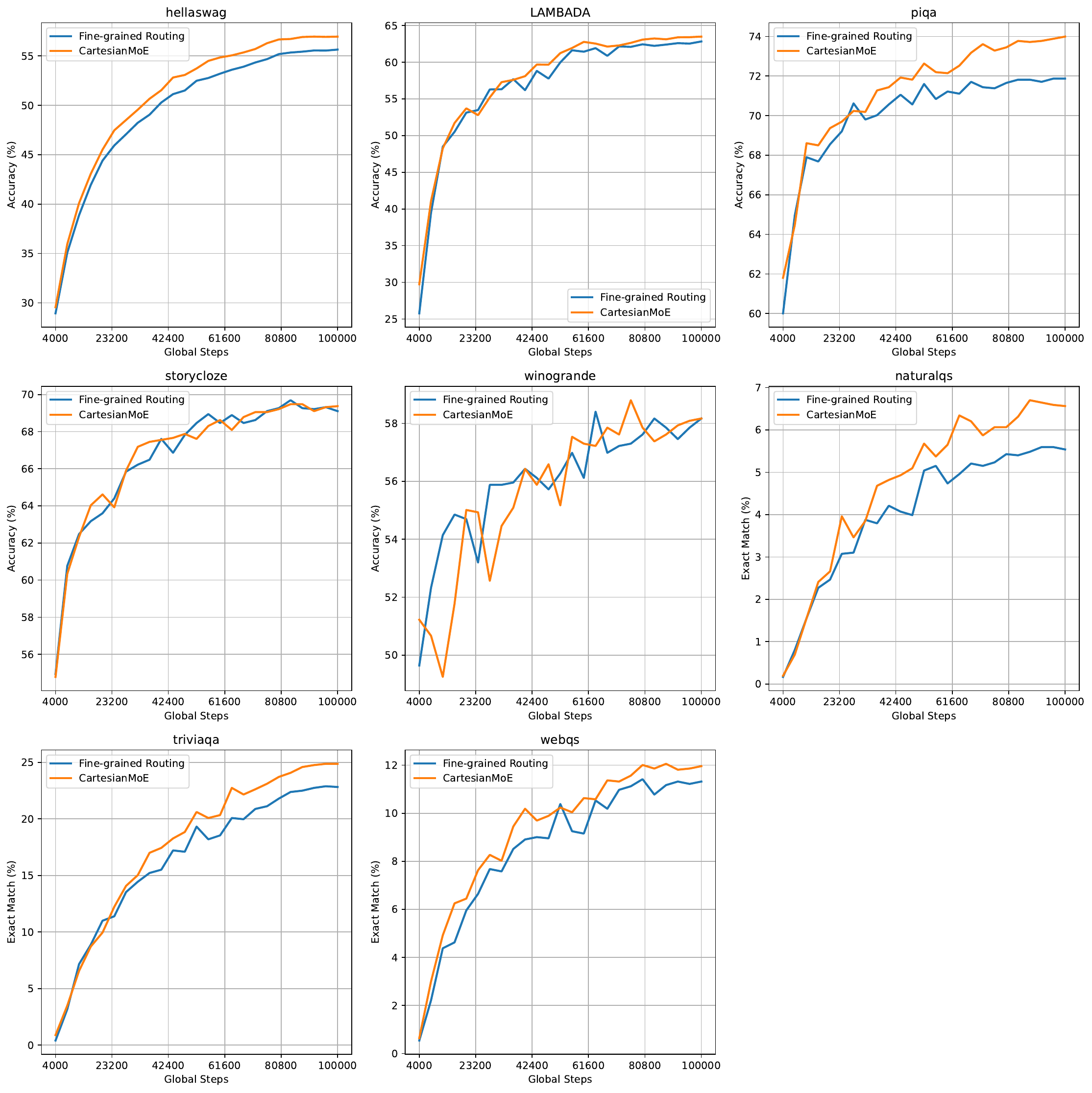}
    \captionof{figure}{Changing curves of downstream task performance during language model training with 400B tokens for CartesianMoE and \textit{Fine-grained Routing} with 7.25B parameters.}
    \label{fig:down_schedule_7B}
\end{figure*}

\end{document}